\pdfoutput=1

\documentclass[11pt]{article}

\usepackage{naacl2021} 

\usepackage{times}
\usepackage{latexsym}
\usepackage[T1]{fontenc}

\usepackage[utf8]{inputenc}

\usepackage{microtype}
\usepackage{xcolor}
\usepackage{float}
\usepackage{url}
\usepackage{multirow}
\usepackage{scalerel,xparse}
\NewDocumentCommand\faceWithTearsOfJoy{}{
    \scalerel*{
        \includegraphics{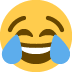}
    }{X}
}
\NewDocumentCommand\clappingHangs{}{
    \scalerel*{
        \includegraphics{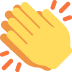}
    }{X}
}
\NewDocumentCommand\flagParaguay{}{
    \scalerel*{
        \includegraphics{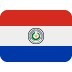}
    }{X}
}

\restylefloat{table}

\title{
On the logistical difficulties and findings of Jopara Sentiment Analysis
}

\author{Marvin M. Ag{\"u}ero-Torales \\
  DECSAI, University of Granada \\ Granada, Spain 
  \\
  \texttt{maguero@correo.ugr.es} \\\And
  David Vilares \\
  Universidade da Coru{\~n}a, CITIC \\ A Coru{\~n}a, Spain 
  \\
  \texttt{david.vilares@udc.es} \\\AND
  Antonio G. L{\'o}pez-Herrera \\
  DECSAI, University of Granada \\ Granada, Spain 
  \\
  \texttt{lopez-herrera@decsai.ugr.es} \\}

\begin{document}
\maketitle
\begin{abstract}
This paper addresses the problem of sentiment analysis for Jopara, a code-switching language between Guarani and Spanish.
We first collect a corpus of Guarani-dominant tweets and discuss on the difficulties of finding quality data for even relatively easy-to-annotate tasks, such as sentiment analysis. Then, we train a set of neural models, including pre-trained language models, and explore whether they perform better than traditional machine learning ones in this low-resource setup. Transformer architectures obtain the best results, despite not considering Guarani during pre-training, but traditional machine learning models perform close due to the low-resource nature of the problem.

\end{abstract}

\section{Introduction}

Indigenous languages have been often marginalized, an issue that is reflected when it comes to design natural language processing (NLP) applications, where they have been barely studied \cite{mager2018challenges}. One of the places where this is greatly noticed is Latin America, where the dominant languages (Spanish and Portuguese) coexist together with hundreds of indigenous languages such as Guarani, Quechua, Nahuatl or Aymara. 

In this context, the Guarani language plays a particular role.
It is an official language in Paraguay and Bolivia. 
Besides, it is spoken in other regions, e.g. Corrientes (Argentina) or Mato Grosso do Sul (Brazil), alongside with their official languages. Overall, it has about 8M speakers.
Its coexistence with other languages, mostly Spanish, has contributed to its use in code-switching setups \cite{muysken1995code,gafaranga2002interactional,matras2020language} and led to Jopara, a code-switching between Guarani and Spanish, with flavours of Portuguese and English \cite{estigarribia2015guarani}.



Despite its official status, there is still few NLP resources developed for Guarani and Jopara. \newcite{abdelali2006guarani} developed a parallel Spanish-English-Guarani corpus for machine translation. Similarly, \newcite{chiruzzo-etal-2020-development} developed a Guarani-Spanish parallel corpus aligned at sentence-level. 
There are also a few online dictionaries and translators
from Guarani to Spanish and other languages.\footnote{\url{https://gn.wiktionary.org/}, \url{https://es.duolingo.com/dictionary/Guarani/}, \url{https://www.paraguay.gov.py/traductor-guarani}, \url{https://www.iguarani.com/}, \url{https://glosbe.com/gn}, and \texttt{Mainumby} \cite{gasser2018mainumby}.}
Beyond machine translation, \newcite{maldonado2016ene} released a corpus for Guarani speech recognition that was collected from the web; and \newcite{rudnick2018cross} presented a system for cross-lingual word sense disambiguation from Spanish to Guarani and Quechua languages. There also are a few resources for PoS-tagging and morphological analysis of Guarani, such as the work by \newcite{uralicnlp_2019} and Apertium;\footnote{\url{https://github.com/apertium/apertium-grn}} and also for parsing, more specifically for the Mby{\'a} Guarani variety \cite{dooley2006lexico, thomas-2019-universal}, under the Universal Dependencies framework.

In the context of sentiment analysis  \citep[SA;][]{pang-etal-2002-thumbs,liu2012sentiment}, and more particularly classifying the polarity of a text as positive, negative or neutral, we are not aware of any previous work; with the exception of \cite{rios2014sentiment}. They presented a sentiment corpus for the Paraguayan Spanish dialect, which also includes words in English and
Portuguese. However, there were few, albeit relevant, words of Guarani ($70$) and Jopara\footnote{Tokens that mix n-grams of characters from Guarani and Spanish, e.g.: `\emph{I understand}' would be `\emph{entiendo}' (es), `\emph{ahechakuaa}' (gn) and `\emph{aentende}' (jopara).} ($10$), in comparison to the amount of the ones in Spanish ($3,802$) \cite[p.~40, Table~II]{rios2014sentiment}.
Overall, SA has focused on rich-resource languages for which data is easy to find, even when it comes to code-switching setups \cite{vilares2016cs}, maybe with a few exceptions such as English code-switched with languages found in India \cite{sitaram2015sentiment,patra2018sentiment,chakravarthi2020corpus}.
In this context, although some previous work has developed multilingual lexicons and methods 
\cite{chen-skiena-2014-building,vilares2017universal}; for languages such as Guarani and other low-resource cases (where web text is scarce), it is hard to develop NLP corpora and systems.

\paragraph{Contribution} Our contribution is twofold. First, we collect a corpus for polarity classification of Jopara tweets, which mixes Guarani and Spanish languages, being the former the dominating language in the corpus. We also discuss on the difficulties that we had to face when creating such resource, such as finding enough Twitter data that shows sentiment and contains a significant amount of Guarani terms.
Second, we train a set of neural encoders and also traditional machine learning models, in order to have a better understand of how old versus new models perform in this low-resource setup, where the amount of data matters.


\section{JOSA: The Jopara Sentiment Analysis dataset}

In what follows, we describe our attempts to collect Jopara tweets. Note that ideally we are interested in tweets that are as Guarani as possible. However, Guarani is intertwined with Spanish, and thus we have focused on Jopara, aiming for Guarani-dominant tweets, in contrast to \citet{rios2014sentiment}. We found interesting to  report failed attempts to collect such data, since the proposed methods would most likely work to collect data in rich resource languages. We hope this can be helpful for other researchers interested in developing datasets for low-resource languages in web environments.

In this line, Twitter does not allow to automatically crawl Guarani tweets, since it is not included in its language identification tool. To overcome this, we considered two alternatives: (i) using a set of Guarani keywords (\S \ref{section-search-keywords}), and (ii) scrapping Twitter accounts that mostly tweet in Guarani (\S \ref{section-search-account}).

\subsection{Downloading tweets using Guarani keywords - An unsuccessful attempt.} \label{section-search-keywords}


As the Twitter real-time streamer can deal with a limited number of keywords, we consider 50 different keywords which are renewed every 3 hours, and used them to sample tweets. To select such keywords, we considered two options:

\begin{enumerate}
    \item\label{method-dictionary-based}  \emph{Dictionary-based keywords}: We used 5.1K Guarani terms from a Spanish-Guarani word-level translator.\footnote{\url{https://github.com/SENATICS/traductor-espanhol-guarani}}
    We then downloaded 2.1M tweets and performed language identification with three tools: (i) \texttt{polyglot},\footnote{\url{https://polyglot.readthedocs.io/en/latest/Detection.html}} (ii) \texttt{fastText} \cite{joulin2016bag} and (iii) \texttt{textcat}.\footnote{\url{https://www.nltk.org/_modules/nltk/classify/textcat.html
    %}, which to identify a language compares the distance measure between language n-grams, when min and max distances are normalized using the \textit{softmax} function to get the confidence in the range of $0$ - $1$.
    }} We assume that the text was Guarani if at least one of them classified the text as Guarani.
    After this, we got 5.3K tweets. Next, a human annotator was in charge of classifying such subset, obtaining that only 150 tweets, over the initial set of 2.1M samples, were prone to be Guarani-dominant.
    
    \item \emph{Corpus-based keywords}: We first merged two Guarani datasets\footnote{BCP-47 \textit{gn} and \textit{gug} codes.} \cite{scannell2007crubadan}, that were generated from web sources and included biblical passages, wiki entries, blog posts or tweets, among other sources. From there, we selected 550 terms, including word uni-grams and bi-grams with 100 occurrences or more.
     Again, we downloaded tweets using the keywords and collected 7M of tweets, but after repeating the language identification phase of step \ref{method-dictionary-based}, we obtained a marginal amount of tweets that were Guarani-dominant.
    
    
\end{enumerate}



\paragraph{Limitations} This approach suffered from a low recall when it came to collect Guarani-dominant tweets, while similar approaches have worked when collecting data for rich-resource languages, where a few keywords were enough to succesfully download tweets in the target language \cite{zampieri-etal-2020-semeval}.
In this context, even if tweets contained a few Guarani terms, there were other issues: (i) words that have the same form in Spanish and Guarani such as `\emph{mano}' (\emph{`hand'} and \emph{'to die}'),
(ii)
loanwords,\footnote{Frequent in Paraguay and border countries \cite{pinta2013lexical}.} such as `\emph{pororo}' (`\emph{popcorn}') or \emph{`chipa'} (traditional Paraguayan food, non-translatable); (iii) or simply tweets where the majority of the content was written in Spanish.
Overall, this has been a problem experienced in other low-resource setups \cite{hong2011language, kreutz-daelemans-2020-streaming}, so we decided instead to look for alternatives to find Guarani-dominant tweets.


\subsection{Downloading tweets from Guarani accounts - A successful attempt.}\label{section-search-account}

In this case, we crawled Twitter accounts that usually tweet in Guarani.\footnote{We followed \url{http://indigenoustweets.com/gn/}. We did not use an external human annotator as in \ref{method-dictionary-based}, since the crawled accounts tend to tweet in Guarani.} We scrapped them, and obtained more than 23K Guarani and Jopara tweets from a few popular users (see Appendix \ref{apx:tw-user-accounts}).
Using the same Guarani language identification approach as in \ref{method-dictionary-based}, we obtained $8,716$ tweets. To eliminate very similar  
tweets that could contaminate the dataset, we removed tweets with a similarity greater than 60\%, according to the Levenshtein distance. After applying this second cleaning step, we obtained a total of 3,948 tweets.



The dataset was then annotated by two native speakers of Guarani and Spanish. They were asked to: (i) determine whether the tweet was strictly written in Guarani, Jopara or other language (i.e., if the tweet did not have any words in Guarani); and determine whether the tweet was positive, neutral or negative. 
For sentiment annotations consolidation, we proceeded similarly to the SemEval-2017  Task 4 guidelines \cite[\S~3.3]{rosenthal2017semeval}.\footnote{We obtained a slight agreement following Cohen's kappa metric \cite{artstein2008inter}.} 
We then filtered the corpus by language, including only those labeled as Guarani or Jopara, to ensure the samples are Guarani-dominant.
This resulted into $3,491$ tweets. 

\paragraph{Limitations} Although this second approach is successful when it comes to collect a reasonable amount of Guarani-dominant tweets, it also suffers from a few limitations. For instance, the first part of Table \ref{tab:sa-stats} shows that due to the nature of the crawled Twitter accounts (who tweet about events, news, announcements, greetings, ephemeris, tweets to encourage the use of Guarani, etc.), there is a tendency to neutral tweets. Also, as the number of selected accounts was small, the number of discussed topics might be limited too. We comment on this a bit further in the Appendix \ref{apx:tw-user-accounts}.

\paragraph{Balanced and unbalanced versions} As we are interested in identifying sentiment in Jopara tweets, we also created a balanced version of JOSA. Note that unbalanced settings are also interesting and might reflect real-world setups. Thus, we will report results both on the unbalanced and balanced setups. More particularly, 
we split each corpus into training (50\%), development (10\%), and test (40\%). We show the statistics in Table \ref{tab:sa-stats}. 

For completeness, in Table \ref{tab:sa-words-corpus} we show for the balanced corpus the top five most frequent terms  (we only consider content tokens) for Guarani, Spanish and some language-independent tokens, such emoticons. This was done based on a manual annotation of a Guarani-Spanish native speaker. 

\begin{table} [htbp]
\footnotesize
\centering
\setlength\tabcolsep{1.5pt}
\begin{tabular}{lrrrr} 
\hline
\textbf{Version} & \textbf{Total} & \textbf{Positive} & \textbf{Neutral} & \textbf{Negative} \\
\hline
Unbalanced & $3,491$ & \multirow{2}{*}{$349$} & $2,728$ & \multirow{2}{*}{414} \\
Balanced & $1,526$ & & $763$ &  \\
\hline
&&&&\\
\hline
 \textbf{Version} & & \textbf{Train} & \textbf{Development} & \textbf{Test} \\
\hline
Unbalanced & $3,491$ & $1,745$ & $349$ & $1,397$  \\
Balanced & $1,526$ & $763$ & $152$ & $611$  \\
\hline
\end{tabular}
\caption{\label{tab:sa-stats}
JOSA statistics and splits for the unbalanced/balanced versions.
}
\vspace{-2mm}
\end{table}

\begin{table} [hbtp]
\footnotesize{
\centering
\setlength\tabcolsep{1.5pt}
\begin{tabular}{p{1.8cm}p{1cm}p{4.5cm}}
\hline
\textbf{Category} & \textbf{\#Terms} & \textbf{Most frequent} \\
\hline
Guarani & $4,336$ & guaranime, ñe'\~{e}, mba'e, guarani, avei \\
Spanish & $1,738$ & paraguay, guaraní, no, es, día
 \\
Other$^{*}$ & $1,440$ &
 
  alcaraz, su, rt, juan, francisco\\

Mixing & $368$ & guaraníme, departamento-pe, castellano-pe, castellanope, twitter-pe
 \\
Emojis & $112$ & \flagParaguay \faceWithTearsOfJoy \clappingHangs xD :) \\
\hline
\multicolumn{3}{p{7.5cm}}{$^{*}$\scriptsize{We include reserved words, proper nouns, acronyms, etc.}}
\end{tabular}
}
\caption{\label{tab:sa-words-corpus}
Frequent terms for the balanced JOSA.
}
\vspace{-2mm}
\end{table}


\section{Models}\label{section-models}

Due to the low-resource setup, we run neural models and pre-trained language models, but also other machine learning models, such as complement naïve Bayes (CNB) and Support Vector Machines (SVMs) \cite{hearst1998support}, since they are less data hungry, and could help shed some light about the real effectiveness of neural models on Jopara texts. In all cases, the selection of the hyperparameters was done over a small grid search based on the dev set. We report the details in the Appendix \ref{apx:hyperparams-search}.

\paragraph{Naïve Bayes and SVMs}


We tokenized the tweets\footnote{We used the TweetTokenizer from the NLTK library.} and represented them as a 1-hot vector of unigrams with a TF-IDF weighting scheme. We used \newcite{scikit-learn} for training.





\paragraph{Neural networks for text classification} We took into account neural networks that process input tweets as a sequence of token vector representations. More particularly, we consider both long short-term memory networks (LSTM) \cite{hochreiter1997long} and convolutional neural networks (CNN) \cite{lecun1995convolutional}, as implemented in NCRF++ \cite{yang2018ncrf++}. Although the former are usually more common in many NLP tasks, the latter have also showed traditionally a good performance on sentiment analysis \cite{kalchbrenner2014convolutional}. 


For the input word embeddings, we tested: (i) randomly initialized word vectors, following an uniform distribution, (ii) and pre-trained non-contextualized representations and more particularly, FastText's word vectors \cite{bojanowski2017enriching} and BPEmb's subword vectors (including the multilingual version, which supports Guarani) \cite{heinzerling2018bpemb}. In both cases, we also concatenate a second word embedding, computed through a char-LSTM (or CNN).

\paragraph{Pre-trained language models}

We also fine-tuned recent contextualized language models on the JOSA training set.
We tested BERT \cite{devlin-etal-2019-bert} including: (i) beto-base-uncased (a Spanish BERT) \cite{CaneteCFP2020}, and (ii) multilingual bert-base-uncased (mBERT-base-uncased, pre-trained on 102 languages).
We also tried more recent variants of multilingual BERT, in particular XLM 
\cite{lample2019cross}. 
Note that BERT models use a wordpiece tokenizer \cite{wu2016google} to generate a vocabulary of the most common subword pieces, rather than the full tokens, and that in the case of the multilingual models, none of the language models used considered Guarani during pre-training.\\

\section{Experiments}

\paragraph{Reproducibility} The baselines and tweet IDs\footnote{Contact the authors for more details.} are available at \url{https://github.com/mmaguero/josa-corpus}.\\

\noindent We run experiments for the unbalanced and balanced versions of JOSA, 
evaluating the macro-accuracy (to mitigate the impact of the neutral class in the unbalanced setup).
Table \ref{tab:sa-results} shows the comparison. Note that all models, even the non-deep-learning models,
only use raw word inputs and do not consider any additional information or hand-crafted features,\footnote{In order to keep an homogeneous evaluation setup.} yet they obtained results that are in line with those of more recent approaches. 

\begin{table} [hbtp]
\footnotesize{
\centering
\setlength\tabcolsep{1.5pt}
\begin{tabular}{lcc}
\hline
 & \multicolumn{2}{p{3.7cm}}{\centering\textbf{Corpus}} \\
\textbf{Model} & \textbf{Unbalanced} & \textbf{Balanced}  \\
\hline
CNB & 0.50 & 0.55  \\
SVM  & 0.55 & 0.54  \\
\hline
$^{C}$CNN-$^{W}$BiLSTM & 0.45 & 0.57  \\
$^{C}$BiLSTM-$^{W}$CNN & 0.49 & 0.53  \\
\hline
$_{BPEmb,gn}$ $^{C}$CNN-$^{W}$BiLSTM & 0.46 & 0.53  \\
$_{BPEmb,gn}$ $^{C}$BiLSTM-$^{W}$CNN & 0.42 & 0.50  \\
\hline
$_{BPEmb,es}$ $^{C}$CNN-$^{W}$BiLSTM & 0.45 & 0.52  \\
$_{BPEmb,es}$ $^{C}$BiLSTM-$^{W}$CNN & 0.45 & 0.50  \\
\hline
$_{BPEmb,m}$ $^{C}$CNN-$^{W}$BiLSTM & 0.47 & 0.52  \\
$_{BPEmb,m}$ $^{C}$BiLSTM-$^{W}$CNN & 0.43 & 0.48 \\
\hline
$_{FastText,gn}$ $^{C}$CNN-$^{W}$BiLSTM & 0.46 & 0.53  \\
$_{FastText,gn}$ $^{C}$BiLSTM-$^{W}$CNN & 0.42 & 0.51  \\
\hline
$_{FastText,es}$ $^{C}$CNN-$^{W}$BiLSTM & 0.46 & 0.52  \\
$_{FastText,es}$ $^{C}$BiLSTM-$^{W}$CNN & 0.46 & 0.46  \\
\hline
BETO$_{base, uncased}$ & \textbf{0.64} & \textbf{0.64}  \\
mBERT$_{base, uncased}$ &0.55 & 0.58  \\
XLM-MLM-TLM-XNLI-15 & 0.46 & 0.49  \\
\hline
\multicolumn{3}{p{7.4cm}}{$^{C}$\scriptsize{Encodes character sequence}. $^{W}$\scriptsize{Encodes word sequence.}} \\
\multicolumn{3}{p{7.4cm}}{\scriptsize{Pre-trained embeddings are represented with a prefix together with their language ISO 639-1 code (except for m: multilingual).}} \\ \end{tabular}
}
\caption{\label{tab:sa-results}
Experimental results on JOSA, both on the balanced and unbalanced setups.
}
\vspace{-2mm}
\end{table}

With respect to the experiments with CNNs and BiLSTMs encoders, we tested different combinations using character representations,
which output is first concatenated to a second external word vector (as explained in \S \ref{section-models}), and then fed to the encoder. Among those, the model that used a character-level CNN and a word-level BiLSTM encoder obtained the best results. Still, the difference with respect to traditional machine learning models is small. We hypothesize this might be due to the low-resource nature of the task. Finally, the pre-trained language models that use transformers architectures, in particular BETO, obtain overall the best results, despite not being pre-trained on Guarani. 
We believe this is partly due to the presence of Spanish words in the corpora and also to the cross-lingual abilities that BERT model might explode, independently of the amount of word overlap \cite{wang2019cross}.



\begin{figure}[hbtp]
\centering
\includegraphics[width=13cm, height=13cm,keepaspectratio]{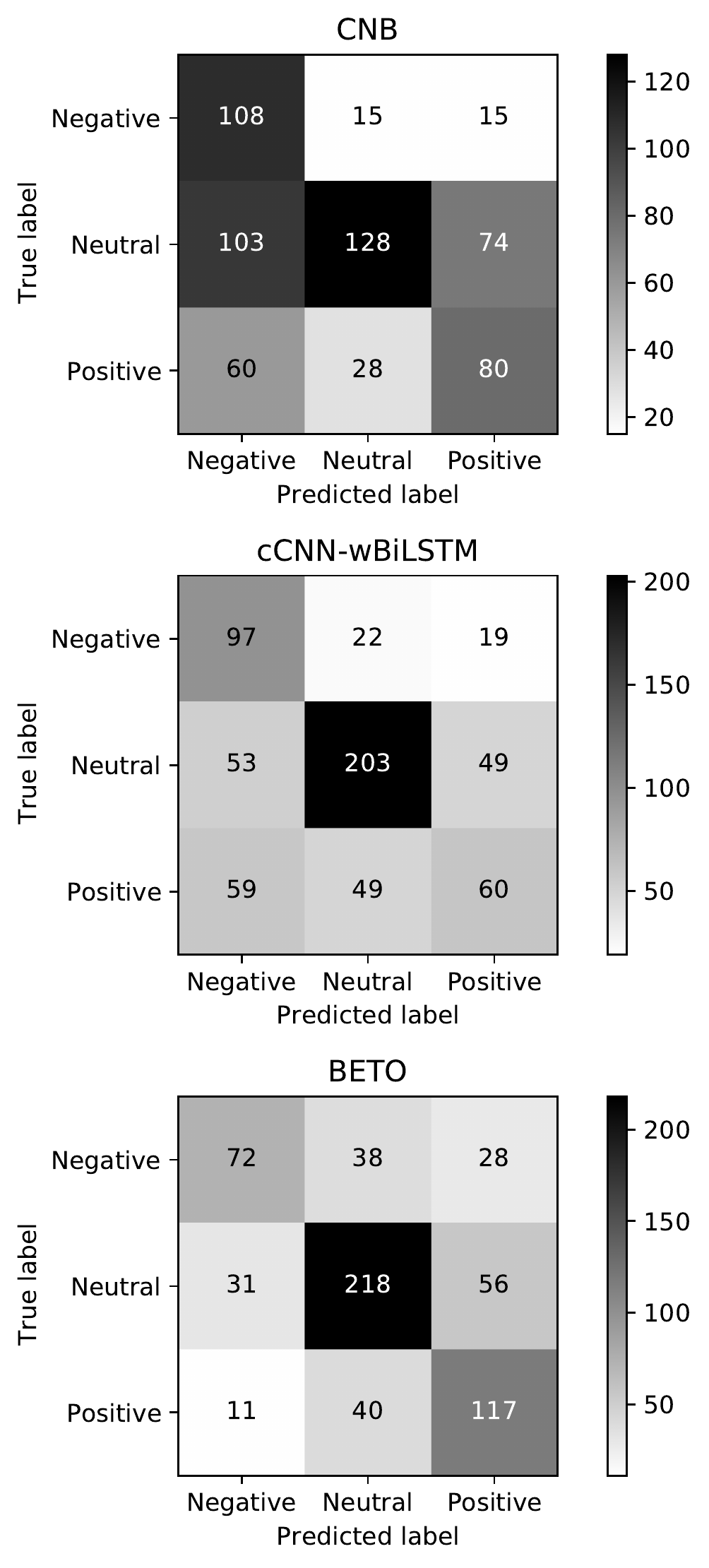}
\caption{Confusion matrix for the balanced version of JOSA and the predictions of a representative member of each machine learning family: CNB, a BiLSTM-based model and Spanish BERT (BETO).}\label{fig:sa-models-cm}
\end{figure}

\paragraph{Error analysis on the balanced version of JOSA} Figure \ref{fig:sa-models-cm} shows the confusion matrices for a representative model of each machine learning family (based on the accuracy): (i) CNB, (ii) the best BiLSTM-based model (CNN-BiLSTM), and (iii) Spanish BERT (BETO). There seems to be different tendencies in the miss-classifications that different models make. For instance, CNB tends to over-classify tweets as negative, while both deep learning models show a more controlled behaviour when predicting this class. Although for the three models neutral tweets seem to be the easiest to identify,  both deep learning models are clearly better at it. Finally, when it comes to identify positive tweets, BETO seems to show the overall best performance. These different tendencies indicate that an ensemble method could be beneficial for low-resource setups such as the ones that JOSA represent, since the models seem to be complementary to certain extent. In this context, we would like to explore this line of work in the future, following previous studies such as \citet{DBLP:journals/corr/abs-1806-04450}, which showed the benefits of combining different machine learning models for Hindi-English code-switching SA.


\section{Conclusion}

This paper explored sentiment analysis on Jopara, a code-switching language that mixes Guarani and Spanish. We collected the first Guarani-dominant dataset for sentiment analysis, and described some of the challenges that we had to face to create a collection where there is a significant number of Guarani terms. We then built several machine learning (naïve Bayes, SVMs) and deep learning models (BiLSTMs, CNNs and BERT-based models) to shed light about how they perform on this particular low-resource setup. Overall, transformers models obtain the best results, even if they did not consider Guarani during pre-training. This poses interesting questions for future work such as how cross-lingual BERT abilities \cite{wang2019cross} can be exploited for this kind of setups, but also how to improve language-specific techniques that can help process low-resource languages efficiently.



\section*{Acknowledgements}
We thank the annotators that labelled JOSA. We also thank ExplosionAI for giving us access to the Prodigy annotation tool\footnote{\url{https://prodi.gy/}} with the Research License.
DV is supported by a 2020 Leonardo Grant for Researchers and Cultural Creators from the FBBVA.\footnote{The BBVA Foundation accepts no responsibility for the opinions, statements and contents included in the project and/or the results thereof, which are entirely the responsibility of the authors.} DV also receives funding from MINECO (ANSWER-ASAP, TIN2017-85160-C2-1-R), from Xunta de Galicia (ED431C 2020/11), from Centro de Investigación de Galicia `CITIC', funded by Xunta de Galicia and the European Union (European Regional Development Fund- Galicia 2014-2020 Program) by grant ED431G 2019/01.
\bibliography{anthology,custom}
\bibliographystyle{acl_natbib}
\clearpage
\appendix

\section{Appendix}
\label{apx:appendix}

\subsection{Twitter user accounts} \label{apx:tw-user-accounts}

We scraped the following Twitter user accounts and mentions: @ndishpy, @chereraugo, @Pontifex\_grn, @lenguaguarani, @enga\_paraguayo, @SPL\_Paraguay, @rubencarlosoje1, as well as some keywords: `\emph{guaranime}', `\emph{avañe'\~{e}me}', `\emph{remiandu}', `\emph{\#marandu}', `\emph{reikuaavéta}', `\emph{hesegua}', `\emph{reheguápe}', `\emph{rejuhúta}'. 

Note that accounts such as @Pontifex\_grn, @SPL\_Paraguay and @lenguaguarani belong to influential people and organizations. For instance, the first belongs to Pope Franscico, the second to the Secretariat of Linguistic Policy of Paraguay, and the third is the account of the General Director of the `Athenaeum of the Guarani Language and Culture'. On the other hand, the terms `\emph{marandu}' (news) and `\emph{remiandu}' (feeling, sense) are related to news, where the first term means `news' or `to report' and the second is the name of a Paraguayan newspaper section\footnote{\url{https://www.abc.com.py/especiales/remiandu/}} that publishes in Guarani.

\subsection{Hyperparameters search and implementation details} \label{apx:hyperparams-search}

To set the machine learning baselines, two standard classifiers were chosen:
a variant of Na\"ive Bayes, Complement Na\"ive Bayes (CNB) \cite{rennie2003tackling} to correct the `severe assumptions' made by the standard Multinomial NB classifier; and Support Vector Machine (SVM) using weighted classes, to mitigate the effect of unbalanced classes. For the CNB, we set \(\alpha=0.1\) and considered only unigrams, except for the balanced version, where the combined use of unigrams and bigrams showed more robust results. To train the SVMs, we tested different values for the kernels: the \texttt{sigmoid} kernel obtained the best results for the unbalanced version of JOSA, and the \texttt{poly} kernel obtained the best results for the balanced version. 

We used the NCRF++ Neural Sequence Labeling Toolkit \cite{yang2018ncrf++} to train our deep learning models and the Hugging face package \cite{wolf-etal-2020-transformers} for the transformer-based models. Table \ref{tab:sa-models-params} shows the hyper-parameters used to train these models, both for the unbalanced and balanced corpus. The pre-trained embeddings used for Spanish, Guarani (and also the multilingual ones) have 300 dimensions. Finally, we trained the CNN and BiLSTM models for $20$ epochs with a batch size of 10, and the transformer-based models were trained for up to 40 epochs relying on early stopping (set to 3). To train the models we used a NVIDIA Tesla T4 GPU with 16GB.

\begin{table} [H]
\footnotesize{
\centering
\setlength\tabcolsep{1.5pt}
\begin{tabular}{ lr }   
\hline
\textbf{Parameter} & \textbf{Options} \\
\hline
\multicolumn{2}{c}{\textbf{Sklearn}} \\
\hline
TF-IDF-Lowercase & [True, False] \\
TF-IDF-n-grams & [(1,1) - (3,3)] \\
SVM-Kernel & [poly, sigmoid, linear, rbf] \\
CNB-alpha & [1.0, 0.1] \\
\hline
\multicolumn{2}{c}{\textbf{NCRF++}} \\ 
\hline
Optimizer & [Adam, AdaGrad, SGD] \\
Avg. batch loss & [True, False] \\
Learning rate & [5e-5 - 0.2] \\
Char hidden dim. & [100, 200, 400, 800] \\
Word hidden dim. & [50, 100, 200] \\
Momentum & [0.0, 0.9, 0.95, 0.99] \\
LSTM Layers & [1, 2] \\
\hline
\multicolumn{2}{c}{\textbf{Hugging Face}} \\ 
\hline
Eval. steps & [200] \\
Eval. strategy & [steps] \\
Disable tqdm & [False] \\
Eval. batch size & [16, 32] \\
Train batch size & [16, 32] \\
Learning rate & [2e-5 - 3e-5$^{*}$] \\
Dropout & [0.1 - 0.6] \\
Epoch & [30 - 40] \\
Weight decay & [0.0 - 0.3] \\
\hline
\multicolumn{2}{p{7cm}}{$^{*}$\scriptsize{Except for the multilingual models, where 5e-5 was necessary to converge.}} \\
\end{tabular}
\caption{\label{tab:sa-models-params}
Hyperparameters for the training of the models, both for the unbalanced and balanced corpus. 
}
}
\vspace{-2mm}
\end{table}


\end{document}